\documentclass{article}
\usepackage{spconf,amsmath,graphicx}
\usepackage{tikz}
\usepackage{cite}

\title{Variational Inference-Based Dropout in Recurrent Neural Networks for Slot Filling in Spoken Language Understanding}
\makeatletter
\def\name#1{\gdef\@name{#1\\}}
\makeatother
\name{\em Jun Qi$^{1}$, Xu Liu$^{2}$, Javier Tejedor$^{2}$}
\address{1.  Electrical and Computer Engineering, Georgia Institute of Technology, Atlanta, USA \\
2. Institute of Industrial Science, The University of Tokyo, Japan	\\
3. Escuela Politecnica Superior, Universidad San Pablo-CEU, CEU Universities, Madrid, Spain \\
}
\begin{document}
%
\maketitle
\begin{abstract}
This paper proposes to generalize the variational recurrent neural network (RNN) with variational inference (VI)-based dropout regularization employed for the long short-term memory (LSTM) cells to more advanced RNN architectures like gated recurrent unit (GRU) and bi-directional LSTM/GRU. The new variational RNNs are employed for slot filling, which is an intriguing but challenging task in spoken language understanding. The experiments on the ATIS dataset suggest that the variational RNNs with the VI-based dropout regularization can significantly improve the naive dropout regularization RNNs-based baseline systems in terms of F-measure. Particularly, the variational RNN with bi-directional LSTM/GRU obtains the best F-measure score.

\end{abstract}
\section{Introduction}
\label{sec:intro}

Slot filling is one of the major but challenging tasks in spoken language understanding because it aims to automatically extract semantic concepts by assigning a set of task-related slots to each word in a sentence. \cite{mesnil2013investigation} was the first reported work that applied recurrent neural network (RNN) to the slot filling task and encouraged the follow-up deep learning work for the task~\cite{qi2020analyzing, qi2020mean, qi2019theory}. The next works focused on deep learning: \cite{hakkani2016multi} tried to replace the vanilla RNNs with more advanced RNN cells based on long short-term memory (LSTM) \cite{hochreiter1997long} or bi-directional LSTM \cite{graves2005framewise}, \cite{guo2014joint} focused on recursive neural networks, and \cite{liu2016attention} utilizes an attention-based RNN. 

In this study, we firstly generalize the variational inference (VI)-based dropout regularization in the LSTM-RNNs to more advanced RNN architectures such as gated recurrent unit (GRU) \cite{cho2014learning} and bi-directional LSTM/GRU. Then, the RNN models with the VI-based dropout regularization are employed in the slot filling task on the ATIS database. Compared with \cite{hakkani2016multi}, this work presents a slight modification of the LSTM-RNNs that can lead to better baseline result, and more RNN architectures with and without VI-based dropout regularization are tested in our experiments. As opposed to \cite{liu2016attention}, our methods are much easier to implement than the attention-based RNN, but similar results can be obtained in practice. 

Since it has been shown that RNNs overfit very quickly \cite{zaremba2014recurrent}, various regularization methods, such as early stopping or small and under-specified models \cite{erhan2010@does}, have to be used during the RNN training stage. Although dropout is normally taken as a simple and effective regularization to overcome the problem of overfitting in deep neural networks \cite{dahl2013improving, qi2020tensor}, it has been concluded that the naive dropout regularization to recurrent weights in RNNs cannot reliably solve the RNN overfitting problem because noise added in the recurrent connections leads to model instabilities \cite{pachitariu2013regularization}. 

However, a recent work \cite{gal2016theoretically} has shown that dropout regularization is a variational approximation technique in Bayesian learning. In addition, the variational inference provides a new variant of dropout regularization, where the same dropout masks are separately shared along time for embedding, decoding, and recurrent weights, so that they can be successfully applied to recurrent layers in RNNs. 

The remainder of the paper is organized as follows: Section $2$ presents the VI-based dropout regularization in RNNs. Section $3$ develops the GRU and bi-directional LSTM/GRU-based RNNs with the VI-based dropout regularization. Section $4$ shows the experimental results on the ATIS database and the paper is concluded in Section $5$. 

\section{Variational Inference-based Dropout Regularization}
\label{sec:vi}

\begin{figure}[htb]
\centering
 \centerline{\includegraphics[width=8.0cm]{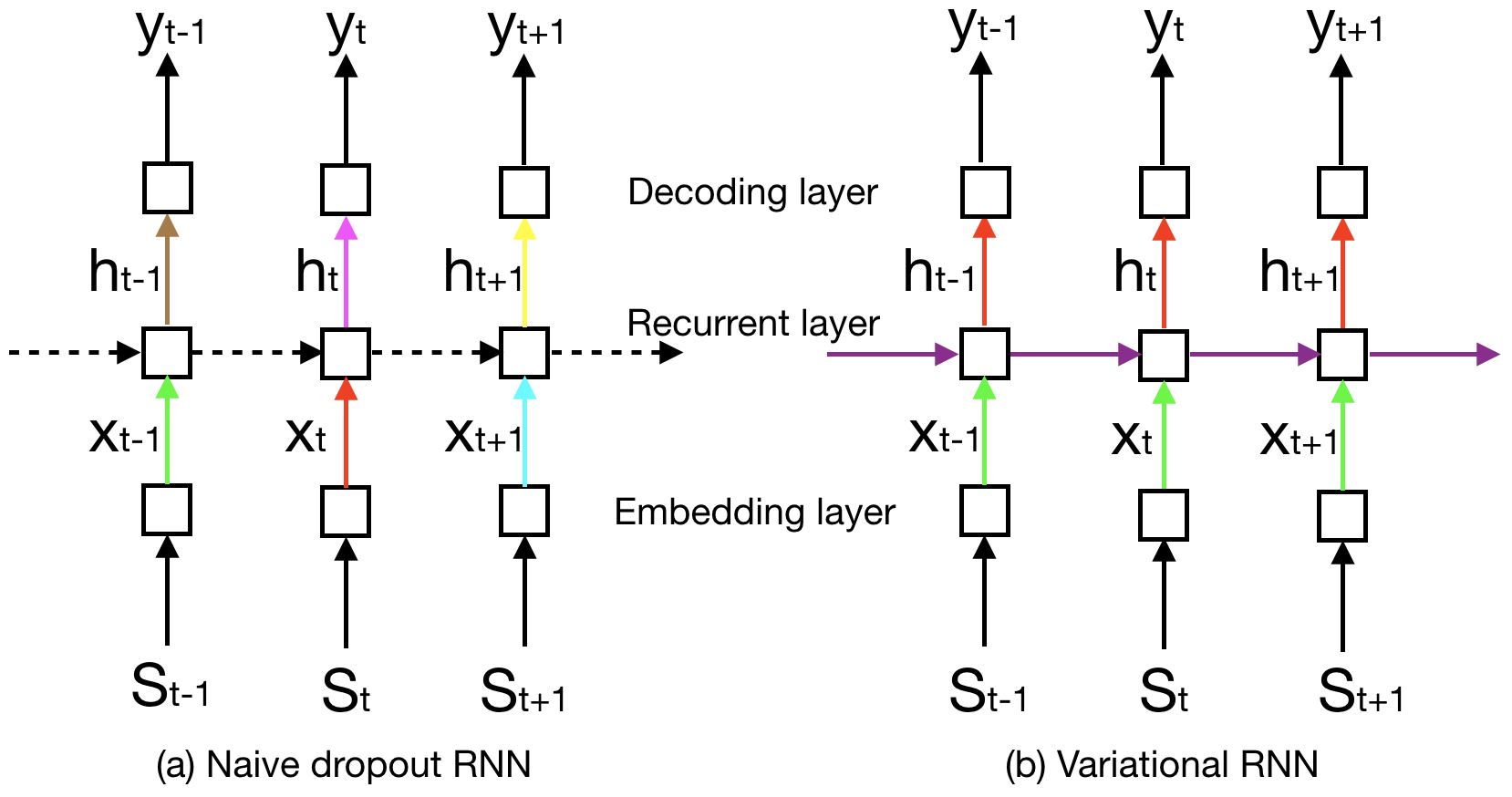}}
 \caption{Dropout regularization in RNNs.}
 \label{fig:dropout_rnn}
\end{figure}

The naive dropout RNN is shown in Figure~\ref{fig:dropout_rnn} (a), and the variational RNN with the VI-based dropout regularization is displayed in Figure~\ref{fig:dropout_rnn} (b). As shown in Figure~\ref{fig:dropout_rnn}, an RNN consists of three layers: the bottom layer is the embedding layer that transforms the $1$-hot representation of every symbolic word ($s_{1}, s_{2}, ..., s_{T}$) into the numerical word embedding ($x_{1}, x_{2}, ..., x_{T}$); the middle layer is the recurrent layer which is evolved over the time and generates outputs of hidden states ($h_{1}, h_{2}, ..., h_{T}$); and the top layer is the decoding layer that outputs the slots ($y_{1}, y_{2}, ..., y_{T}$) associated with the symbolic words. The colored arrows denote different dropout masks applied to the weights in the layers, and the same colored arrows mean the same dropout masks are used. 

In the framework of naive dropout RNN, different dropout masks are applied to both embedding and decoding layers rather than the recurrent layer. However, the variational RNN does not only allow the use of dropout regularization in the recurrent layer, but the same three dropout masks are also separately shared in the embedding, recurrent, and decoding layers.  

Next, the VI-based dropout regularization is briefly summarized. Given a sequence of input vectors $X = \{x_{1}, ..., x_{T}\}$ associated with the label set $Y= \{y_{1}, y_{2}, ..., y_{T}\}$, a set of model weight matrices is defined as a random variable $\omega = \{W, U, V, b\}$, where $W, U, V$ separately denote weights of embedding, recurrent, and decoding layers, respectively, and $b$ refers to the bias for the recurrent layer. In addition, $\omega$ has a prior $p(\omega)$ that is assumed to be a Gaussian distribution. Furthermore, the activation function of an RNN unit is defined as $h_{t} = f_{h}^{\omega}(x_{t}, h_{t-1})$ and the output of the decoding layer is $f^{\omega}(x) = f_{y}^{\omega}(h_{T})$. A typical function $f_{h}^{\omega}(x_{t}, h_{t-1})$ is $tanh(\cdot)$ as defined in (\ref{eq:tanh}).
\begin{equation}
\label{eq:tanh}
f_{h}^{\omega}(x_{t}, h_{t-1}) = tanh(Wx_{t} + Uh_{t-1} + b)
\end{equation}

The variational interpretation of dropout regularization comes from the Kullback-Leibler (KL) divergence as shown in (\ref{eq:kl}), where $p(\omega | X, Y)$ is the true distribution, $N$ is the number of data, and $q(\omega)$ is the approximated distribution that factorizes over weight columns $\omega_{ik}$ through (\ref{eq:delta}), where $m_{k}$ is the variational mean parameter, $p$ is the dropout probability given in advance, and $\sigma^{2}$ is a small variance. 
\begin{equation}
\label{eq:kl}
\begin{aligned}
&KL(q(\omega) || p(\omega | X, Y)) 	\\
\propto& -\int q(\omega) log p(Y | X, \omega) d\omega + KL(q(\omega) || p(\omega)) 	\\
=& -\sum\limits_{i=1}^{N} \int q(\omega) log p(y_{i} | f^{\omega}(x_{i})) d\omega + KL(q(\omega) || p(\omega))
\end{aligned}
\end{equation}
\begin{equation}
\label{eq:delta}
q(\omega_{ik}) = pN(\omega_{ik}; 0, \sigma^{2}) + (1-p)N(\omega_{ik}; m_{k}, \sigma^{2})
\end{equation}
By the factorization of $f^{\omega}(x)$ and applying the Markov Chain integration with $\hat{\omega}_{i} \sim q(\omega)$, the equation (\ref{eq:fact_eq}) holds. 
\begin{equation}
\label{eq:fact_eq}
\begin{aligned}
&\int q(\omega) logp(y_{i} | f^{\omega}(x)) d\omega \\
=&\int q(\omega) log p(y_{i} | f^{\omega}_{y}(f_{h}^{\omega}(x_{iT}, ... f_{h}^{\omega}(x_{i1}, h_{0})...))) d\omega	\\
\approx & log p(y_{i} | f_{y}^{\hat{\omega}_{i}}(f_{h}^{\hat{\omega}_{i}}(x_{iT}, ...f_{h}^{\hat{\omega}_{i}}(x_{i1}, h_{0})...)))
\end{aligned}
\end{equation}
The objective function for the dropout regularization can be derived as (\ref{eg:approx1}), which means that random masks are repeatedly used to set weight columns to zero at each time step for weight matrices of embedding, recurrent, and decoding layers. 
\begin{equation}
\label{eg:approx1}
\begin{aligned}
L_{VI} \approx& -\sum\limits_{i=1}^{N} log p(y_{i} | f_{y}^{\hat{\omega}_{i}}(f_{h}^{\hat{\omega}_{i}}(x_{iT}, ...f_{h}^{\hat{\omega}_{i}}(x_{i1}, h_{0})...))) \\
&+ KL(q(\omega) || p(\omega))
\end{aligned}
\end{equation}

\section{The Implementation of VI-based Dropout Regularization in RNNs}
The previous work~\cite{gal2016theoretically} showed how to apply the VI-based dropout regularization in LSTM-RNNs. On the other hand, this work further generalizes it to more complex RNN architectures such as GRU, and bi-directional RNN with LSTM or GRU cells, as is described next. 

\subsection{RNN with LSTM cells}
It has been claimed that the back-propagation algorithm when used for the RNN training can result in exploding or vanishing gradients~\cite{hochreiter2001gradient}. Although exploding gradients could be alleviated by gradient clipping, this cannot be employed to deal with vanishing gradients. To mitigate the vanishing gradient issue, LSTM cells were thorough designed by introducing a memory vector $c_{t}$ and four mechanism gates $f_{t}$, $i_{t}$, $g_{t}$, and $o_{t}$. 
\begin{figure}[htb]
\centering
 \centerline{\includegraphics[width=8.8cm]{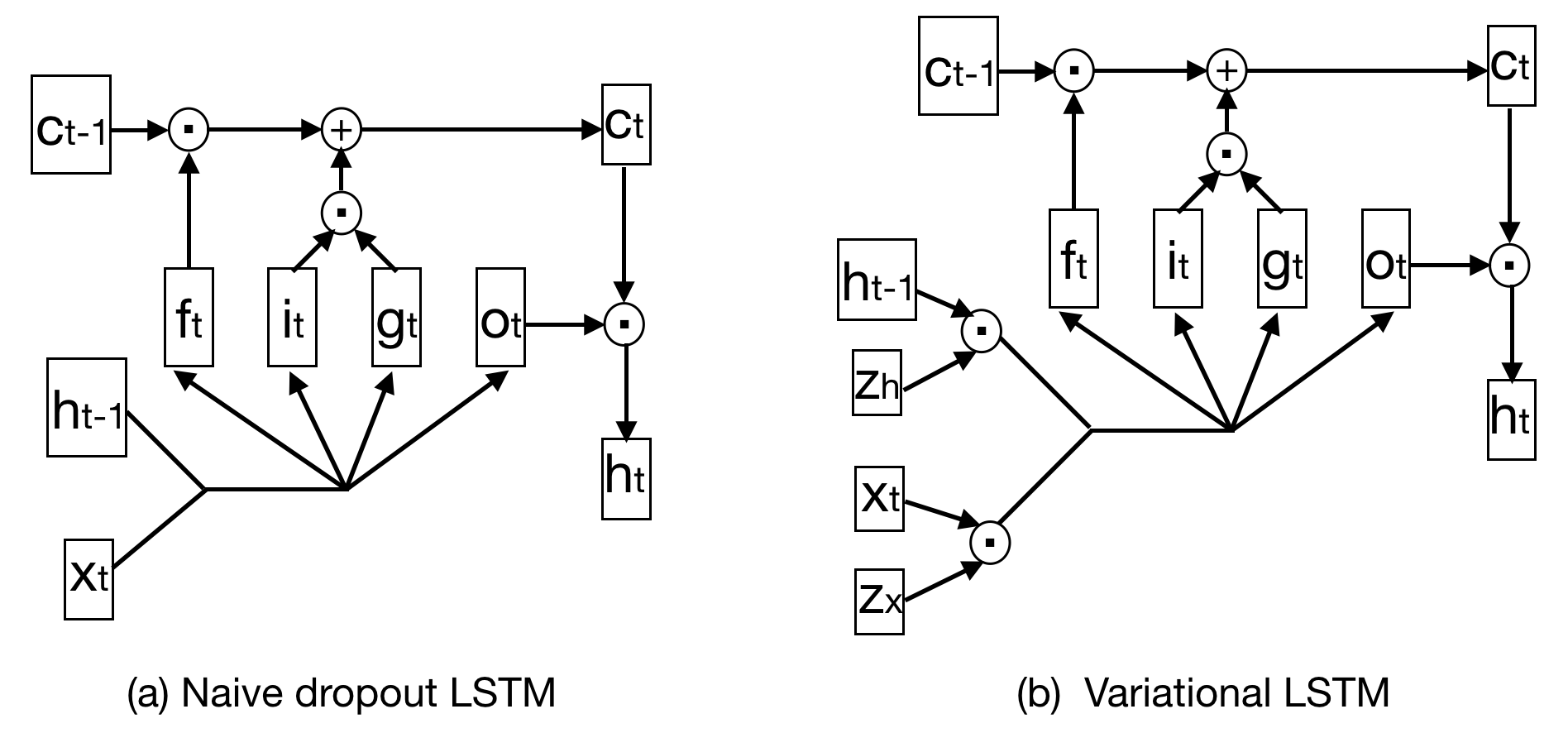}}
 \caption{Dropout regularization in LSTM for RNNs.}
 \label{fig:lstm}
\end{figure}

Figure~\ref{fig:lstm} (a) illustrates the architecture of an LSTM, where the input gates $i_{t}$ and $g_{t}$ scale down the input $x_{t}$, the forget gate $f_{t}$ is used to scale down the memory vector $c_{t}$, and the output gate $o_{t}$ is used to scale down the output before reaching the final $h_{t}$. The mathematical formulations for the LSTM gates are separately shown in (\ref{eq:lstm_1}), (\ref{eq:lstm_2}), (\ref{eq:lstm_3}), and (\ref{eq:lstm_4}), where ($W_{xi}$, $W_{hi}$), ($W_{xf}$, $W_{hf}$), ($W_{xo}$, $W_{ho}$), and ($W_{xg}$, $W_{hg}$) represent weights of gates $i_{t}$, $f_{t}$, $o_{t}$, and $g_{t}$, respectively. The memory vector $c_{t}$ and the hidden state vector $h_{t}$ are shown in (\ref{eq:ct}) and (\ref{eq:ht}), respectively. 
\begin{equation}
\label{eq:lstm_1}
i_{t} = sigm(W_{xi}x_{t} + W_{hi}h_{t-1})
\end{equation}
\begin{equation}
\label{eq:lstm_2}
f_{t} = sigm(W_{xf}x_{t} + W_{hf}h_{t-1})
\end{equation}
\begin{equation}
\label{eq:lstm_3}
o_{t} = sigm(W_{xo}x_{t} + W_{ho}h_{t-1})
\end{equation}
\begin{equation}
\label{eq:lstm_4}
g_{t} = tanh(W_{xg}x_{t} + W_{hg}h_{t-1})
\end{equation}
\begin{equation}
\label{eq:ct}
c_{t} = f_{t} \odot c_{t-1} + i_{t} \odot g_{t}
\end{equation}
\begin{equation}
\label{eq:ht}
h_{t} = o\odot tanh(c_{t})
\end{equation}
As shown in Figure~\ref{fig:lstm} (b), the implementation of variational LSTM just offers two dropout masks $z_{x}$ and $z_{h}$ to $x_{t}$ and $h_{t-1}$, respectively, and the corresponding equations for the LSTM gates are modified as (\ref{eq:lstm_1_m}), (\ref{eq:lstm_2_m}), (\ref{eq:lstm_3_m}), and (\ref{eq:lstm_4_m}).
\begin{equation}
\label{eq:lstm_1_m}
i_{t} = sigm(W_{xi}(x_{t}\odot z_{x}) + W_{hi}(h_{t-1}\odot z_{h}))
\end{equation}
\begin{equation}
\label{eq:lstm_2_m}
f_{t} = sigm(W_{xf}(x_{t}\odot z_{x}) + W_{hf}(h_{t-1}\odot z_{h}))
\end{equation}
\begin{equation}
\label{eq:lstm_3_m}
o_{t} = sigm(W_{xo}(x_{t}\odot z_{x})  + W_{ho}(h_{t-1}\odot z_{h}))
\end{equation}
\begin{equation}
\label{eq:lstm_4_m}
g_{t} = tanh(W_{xg}(x_{t}\odot z_{x}) + W_{hg}(h_{t-1}\odot z_{h}))
\end{equation}

\subsection{RNN with GRU cells}
\begin{figure}[htb]
\centering
 \centerline{\includegraphics[width=8.8cm]{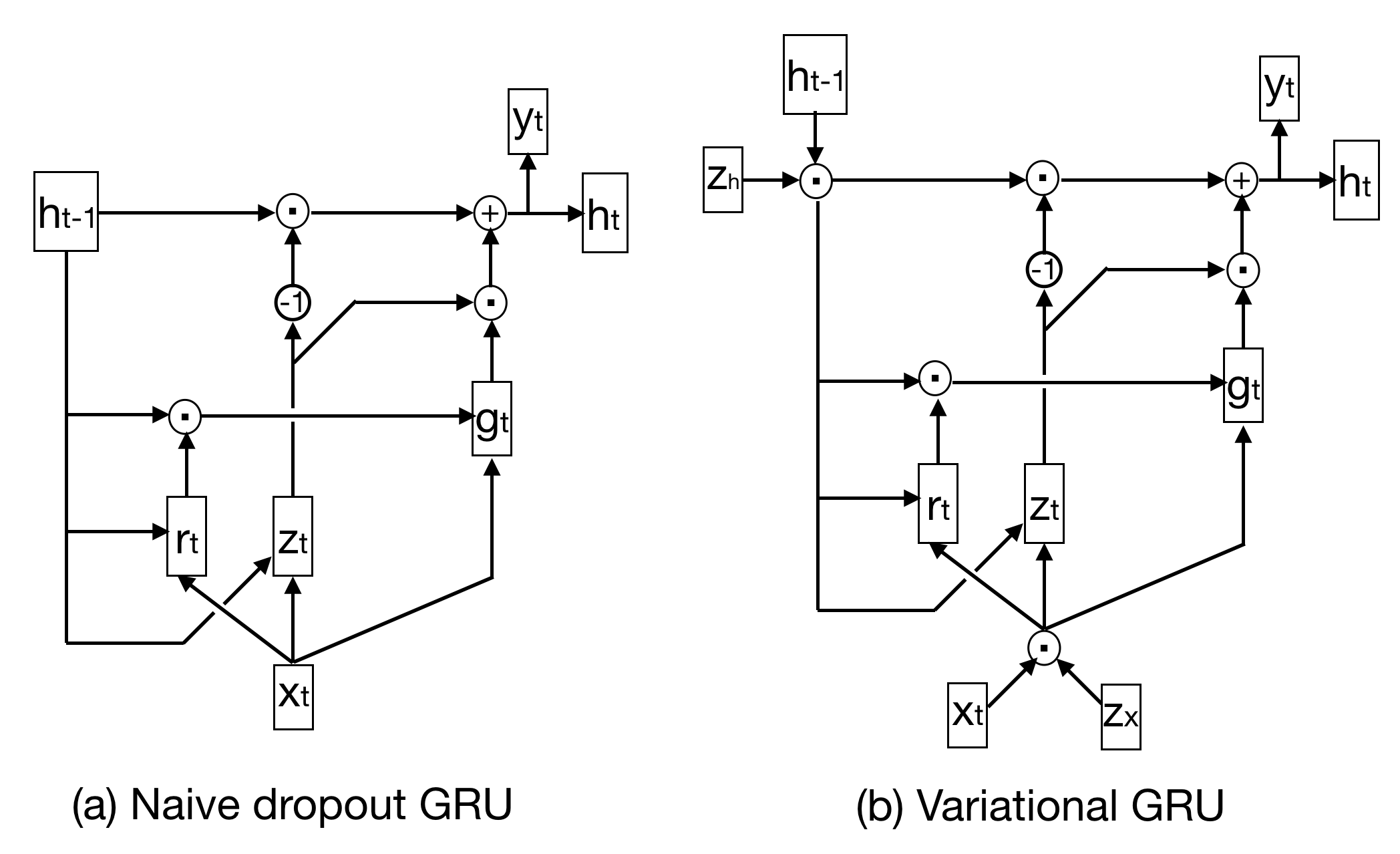}}
 \caption{Dropout regularization in GRU for RNNs.}
 \label{fig:gru}
\end{figure}

GRU is a simplified version of the LSTM cell and normally obtains better results with a lower computational cost. As shown in Figure~\ref{fig:gru} (a), the main differences between GRU and LSTM cells are these: there is no additional memory state vector and all state vectors are merged into a single vector $h_{t}$; a single gate controller $g_{t}$ controls both the forget gate and the input gate; there is no output gate. The mathematical formulation of the GRU cell is shown from (\ref{eq:gru_1}) to (\ref{eq:gru_4}), where $(W_{xz}, W_{hz})$, $(W_{xr}, W_{hr})$, and $(W_{xg}, W_{hg})$ represent weights for the gates $z_{t}$, $r_{t}$, and $g_{t}$, respectively. 
\begin{equation}
\label{eq:gru_1}
z_{t} = sigm(W_{xz}x_{t} + W_{hz} h_{t-1})
\end{equation}
\begin{equation}
\label{eq:gru_2}
r_{t} = sigm(W_{xr} x_{t} + W_{hr} h_{t-1})
\end{equation}
\begin{equation}
\label{eq:gru_3}
g_{t} = tanh(W_{xg} x_{t} + W_{hg} (r_{t} \odot h_{t-1}))
\end{equation}
\begin{equation}
\label{eq:gru_4}
h_{t} = (1 - z_{t}) \odot tanh(W_{xg} h_{t-1} + z_{t} \odot g_{t})
\end{equation}
The implementation of variational GRU causes the modifications of equations (\ref{eq:gru_1}), (\ref{eq:gru_2}), and (\ref{eq:gru_3}) into equations (\ref{eq:gru_5}), (\ref{eq:gru_6}), and (\ref{eq:gru_7}), respectively. 
\begin{equation}
\label{eq:gru_5}
z_{t} = sigm(W_{xz} (x_{t} \odot z_{x}) + W_{hz}(h_{t-1} \odot z_{h}))
\end{equation}
\begin{equation}
\label{eq:gru_6}
r_{t} = sigm(W_{xr} (x_{t}\odot z_{x}) + W_{hr} (h_{t-1} \odot z_{h}))
\end{equation}
\begin{equation}
\label{eq:gru_7}
g_{t} = tanh(W_{xg} (x_{t}\odot z_{x})+ W_{hg} (r_{t} \odot h_{t-1}))
\end{equation}

\subsection{Bi-directional RNNs}
\begin{figure}[htb]
\centering
 \centerline{\includegraphics[width=8.8cm]{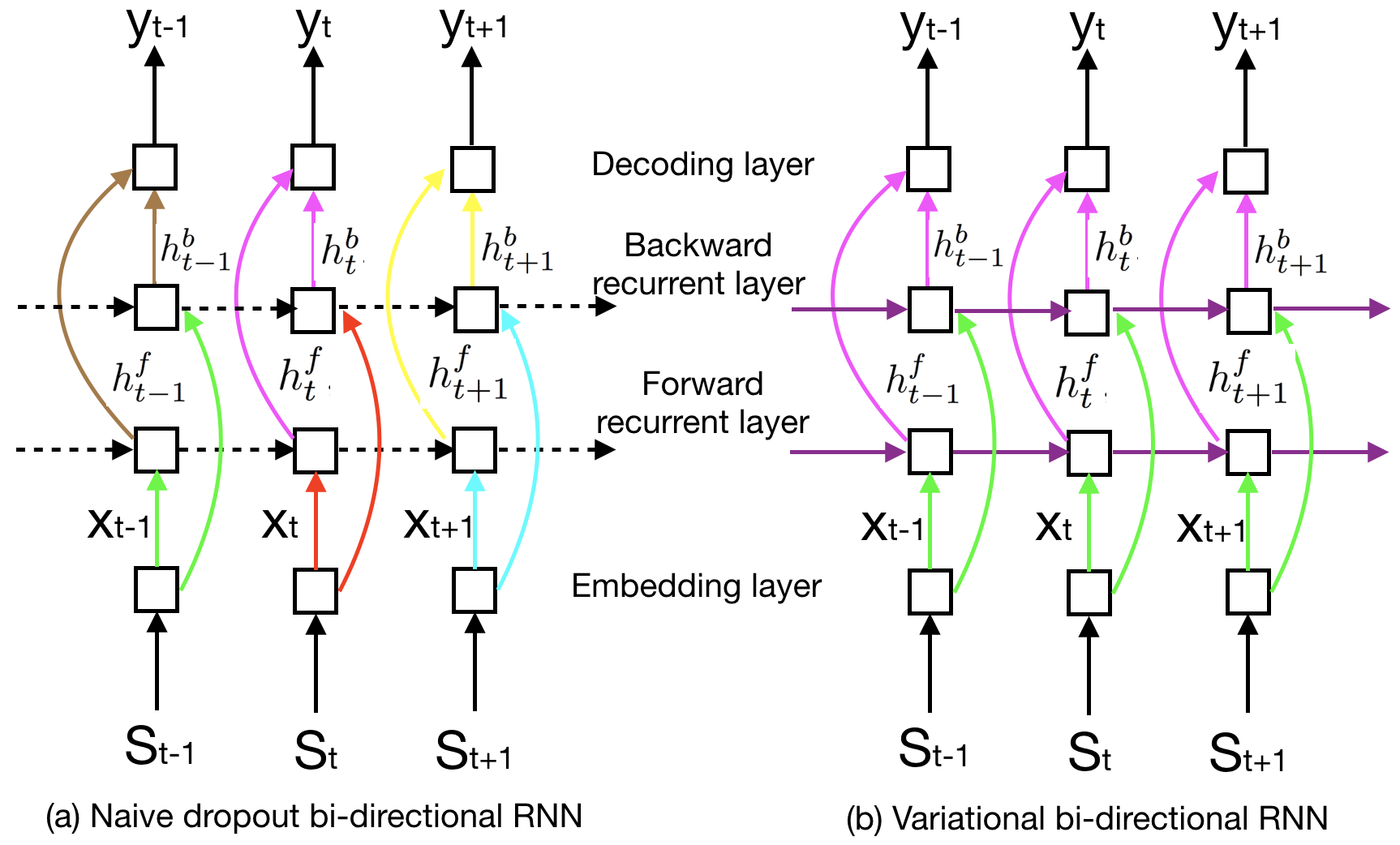}}
 \caption{Dropout regularization in bi-directional RNNs.}
 \label{fig:birnn}
\end{figure}

Bi-directional RNNs can also be implemented with variational LSTM and GRU cells. Figure~\ref{fig:birnn} (a) shows an illustration of two bi-directional RNNs, where two RNN architectures are traversed in left-to-right and right-to-left manners, and their hidden layers are concatenated when computing the output sequence. In addition, naive dropout regularization is applied to the embedding and decoding layers but not to forward nor backward recurrent layers. Note that the same colored arrows denote the use of a same dropout mask. 

Variational bi-directional RNN in Figure~\ref{fig:birnn} (b) implies that the dropout masks are shared through time for embedding, decoding, and the two recurrent layers. Apart from the vanilla RNN, LSTM and GRU can also be used as the \emph{improved} RNN cell in the variational bi-directional RNN architecture.

\section{Experiments}
\subsection{Experimental setup}
Our experiments are conducted on the Airline Travel Information System (ATIS) dataset, which is commonly used for the slot filling task by the spoken language understanding community. The training/validation set contains $4978$ utterances selected from Class A (context independent) training data in the ATIS-3 corpus, which was further divided into $80\%$ of data for training and $20\%$ of data for validation, while the test set contains $893$ utterances from the ATIS-3 Nov$93$ and Dec$94$ datasets. There is a total of $56590$ words in the training/validation set, and $9198$ words in the test set. The average length of the sentences is $15$, and the number of classes (different slots) is $128$.

\begin{figure}[htb]
\centering
 \centerline{\includegraphics[width=8.8cm]{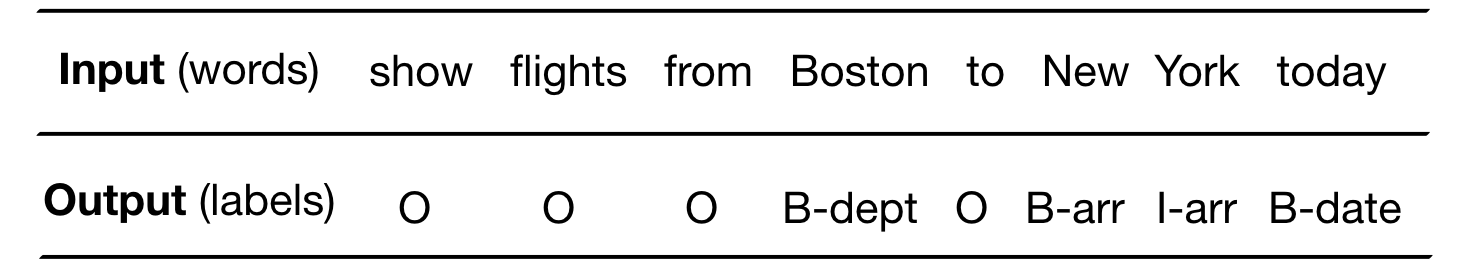}}
 \caption{An illustration of slot filling.}
 \label{fig:slots}
\end{figure}

Figure~\ref{fig:slots} demonstrates an example of slot filling for each word in one utterance, where label O denotes NULL, and B-dept, B-arr, I-arr, and B-date are valid slots for words. 

Since the RNN inputs are represented by the 1-hot representation and are further transformed to $100$-dimensional word embeddings via the embedding layer, the dimension of the inputs corresponds to the size of the vocabulary and the embedding layer involves $100$ units \cite{qi2020submodular, qi2018distributed}. The numbers of units in all recurrent layers for the LSTM and GRU cells are fixed to $100$, and the number of units in the final decoding layer is set to $128$, which corresponds to the number of labels. 

Besides, the activation probability for creating the dropout masks is set to $0.5$ \cite{qi2013bottleneck}. The naive dropout regularization is applied to the embedding and decoding layers in the RNNs with the LSTM, GRU, and bi-directional LSTM/GRU cells, whereas the variational RNNs involve the application of the VI-based dropout regularization for all RNN layers. 

\subsection{Experimental results}
Table $1$ presents the F-measure \cite{qi2016deep} results for different RNN architectures. The \emph{best F} column denotes the best F-measure score obtained by among $10$ different random initializations of the weights of the RNN models introduced in this paper, and the \emph{average F} column refers to the average F-measure score of $10$ runs.  
\begin{table}[htbp]
\begin{center}
\label{tab:1}
\begin{tabular}{c||c|c}
\hline
\textbf{Model}	&\textbf{best F} &\textbf{average F} \\
\hline
LSTM	 &93.08 		&92.42 \\
\hline
GRU		 &94.16	        &93.49 \\
\hline
Bi-directional LSTM	 &95.14	        &94.63 \\
\hline
Bi-directional GRU	&95.32		&94.57 \\
\hline
\hline
Variational LSTM &94.35	 &93.87  \\
\hline
Variational GRU &94.48    &94.15  \\
\hline
Variational bi-directional LSTM &95.55   &94.97 \\
\hline
Variational bi-directional GRU &95.61    &95.04 \\
\hline
\end{tabular}
\caption{F-measure results using ATIS data.}
\end{center}
\end{table}

The results in Table $1$ suggest that the variational RNNs improve their counterpart naive dropout regularization-based RNN model. In addition, the variational bi-directional LSTM and variational bi-directional GRU perform better than the two other variational RNN models. 

Table $2$ compares the results obtained by our methods with previous work on the slot filling task. This includes attention bi-directional RNN~\cite{liu2016attention}, attention encoder-decoder NN~\cite{liu2016attention}, and look-around LSTM (LSTM-LA)~\cite{hakkani2016multi}. 
\begin{table}[htbp]
\begin{center}
\label{tab:2}
\begin{tabular}{c||c}
\hline
\textbf{Model}	&\textbf{F Score}	\\
\hline
LSTM-LA					& 95.32		\\
\hline
Attention encoder-decoder NN &95.78  		\\
\hline
Attention bi-directional RNN			 &95.75  		 \\
\hline
\hline
Variational bi-directional LSTM &95.55  	 \\
\hline
Variational bi-directional GRU &95.61     	\\
\hline
\end{tabular}
\caption{Comparison of F-scores from different RNN models.}
\end{center}
\end{table}

The F-measure scores in Table $2$ show that our variational RNN models are still below the best results obtained by the attention encoder-decoder NN and the attention bi-directional RNN, although those perform better than LSTM-LA. However, the variational RNNs involve simpler neural network architectures with less number of parameters than the two attention-based models, with no significant performance degradation. 

\section{Conclusions}
This work has proposed variational inference-based dropout regularization for RNNs with LSTM, GRU, and bi-directional LSTM/GRU cells. Contrary to the naive dropout regularization for embedding and decoding layers, the VI-based dropout regularization is applied to all RNN layers including recurrent layers by sharing the same dropout masks in the RNN layers. The experiments on the slot filling task on ATIS database showed that the variational RNN models obtain better results than the naive dropout regularization-based RNN models. In particular, the variational bi-directional LSTM/GRU obtains the best results in terms of F-measure. 

\vfill\pagebreak

\label{sec:refs}

\bibliographystyle{IEEEbib}
\bibliography{strings,refs}

\end{document}